\documentclass[letterpaper]{article}
\usepackage{aaai}
\usepackage{times}
\usepackage{helvet}
\usepackage{courier}
\usepackage{graphicx}
\usepackage{hyperref}
\usepackage{amsmath}
\begin{document}
%
\title{Towards AI-driven Integrative Emissions Monitoring \& Management for Nature-Based Climate Solutions}

\author{Olamide Oladeji \textsuperscript{\rm 1,2}, Seyed Shahabeddin Mousavi \textsuperscript{\rm 1,2}\\
    \textsuperscript{\rm 1}Department of Management Science \& Engineering, Stanford University, Stanford, CA, USA\\
    \textsuperscript{\rm 2}Sustainable Finance Initiative, Precourt Institute for Energy, Stanford University, Stanford, CA, USA\\
    \href{mailto:oladeji@stanford.edu}{oladeji@stanford.edu}, \href{mailto:ssmousav@cs.stanford.edu}{ssmousav@cs.stanford.edu}
}

\maketitle
\begin{abstract}
\begin{quote}
AI has been proposed as an important tool to support several efforts related to nature-based climate solutions such as the detection of wildfires that affect forests and vegetation-based offsets. While this and other use-cases provide important demonstrative value of the power of AI in climate change mitigation, such efforts have typically been undertaken in silos, without awareness of the integrative nature of real-world climate policy-making. 
In this paper, we propose a novel overarching framework for AI-aided integrated and comprehensive decision support for various aspects of nature-based climate decision-making. Focusing on vegetation-based solutions such as forests, we demonstrate how different AI-aided decision support models such as AI-aided wildfire detection, AI-aided vegetation carbon stock assessment, reversal risk mitigation, and disaster response planning can be integrated into a comprehensive framework. 
Rather than being disparate elements, we posit that the exchange of data and analytical results across elements of the framework, and careful mitigation of uncertainty propagation will provide tremendous value relative to the status-quo for real-world climate policy-making.

\end{quote}
\end{abstract}

\noindent 

\section{Introduction and Motivation}

Robust accounting of carbon emissions is a critical step in the ongoing efforts to mitigate climate change. In the context of climate change mitigation, carbon accounting involves properly quantifying the amount of greenhouse gas emissions, particularly carbon dioxide (CO2), that are released or sequestered by different entities, ranging from individual projects or companies to entire countries or regions. Nature-based solutions have been a prominent player in the space and, if precise enough so as to result in long-term trust in their viability and reliability in the face of a changing climate with wildfires scorching the earth with ever-growing ferocity and magnitude becoming an ever more frequent occurrence, can be of particular importance as they can help cool the planet by managing, protecting, and restoring ecosystems. These solutions can have a powerful role in reducing temperatures in the long term as land-use changes will continue to act long past the point at which net-zero emissions are achieved and global temperatures peak (known as peak warming), and will have an important role in planetary cooling in the second half of this century. As such, nature-based solutions must be designed for longevity, paying closer attention to their long-term carbon-sink potential, as well as their impacts on biodiversity, equity, and sustainable development goals. \cite{girardin2021nature}

Proferring solutions to mitigate current gaps associated with carbon accounting requires that we adequately characterize the issues. Undeniably, effective carbon accounting is predicated on precise quantification of both carbon sources and sinks within a given system. In the context of forest ecosystems, carbon sequestration is a well-studied phenomenon that plays a crucial role in offsetting anthropogenic emissions. However, natural disturbances, such as wildfires, pose significant challenges to the robustness and reliability of this carbon accounting, largely due to their sporadic nature and the considerable carbon emissions they cause \cite{kurz2008mountain}. 

Overall, there are various dynamics at play when dealing with vegetation-based solutions leading to a complex set of decisionmaking by various agents. For example, in such a complex interacting system, one element of decisionmaking centers around the detection of wildfires and finding ways to automate such a process, the quantification of the emissions impact of wildfires, or even the integration of wildfire impacts and reversal risk in the insurance of vegetation offsets. In the absence of robust systems and computational frameworks to adequately integrate these disparate, often-siloed decision elements, climate mitigation decision-making efforts around nature-based climate solutions remain incomplete and potentially misleading, failing to capture the full spectrum of dynamics. 
With recent advancements in Artificial Intelligence (AI), automation of this process looks evermore promising. AI techniques such as computer vision methods can help provide data and support automation aspects e.g. aiding with the detection and quantification of wildfire disturbances on nature-based climate solutions. Such a use-case means they allow us to appropriately integrate them when undertaking carbon accounting for these solutions. However, AI can also support other decisions connected to this. In this paper, we explore the use of AI for integrative emissions monitoring \& management for nature-based carbon accounting and discuss how such an approach can significantly drive comprehensive decision-making across all aspects of nature-based climate solutions.

\section{Methodology: Developing the AI-driven Integrative Framework}
AI techniques have been used to address various issues affecting nature-based climate solutions such as vegetation-based offset projects. For example, when vegetation projects or natural forests are affected by wildfires, satellite imagery-based computer vision can be used to detect wildfires in real-time, supporting critical disaster response efforts. AI has also been used to improve rapid, remote quantification of carbon stock \cite{xu2020product} \cite{zhang2022estimation}. However, these and other analytical efforts that attempt to leverage AI in nature-based solutions decision-making are typically silo-ed and are often undertaken without recognition of the interlinked nature in real-world policy-making. For example, while policymakers and planners have to detect and plan response to wildfires, they also have to design incentives to mitigate the reversal risk that increases with wildfires. Similarly, efforts to improve carbon accounting practices must be linked with monitoring to be effective. In this section, we present an integrated framework that addresses this gap, delving first into the different elements of the framework.

\subsection{AI-aided Vegetation Wildfire Detection}
Recent advances in deep learning and computer vision, coupled with the increased proliferation of frequent satellite mapping across the world can be leveraged in order to address the challenges of wildfire detection and mapping. For example, in \cite{wu2020satellite}, the authors propose and develop a deep learning wildfire detection system which is trained on images of historical wildfire locations across the United States using the Sentinel-2 satellite. Based on the gaps identified in the literature of deep learning for wildfire detection, \cite{wu2020satellite} explores the use of several image layers obtainable from Sentinel-2’s satellite imagery service. In particular, they focus on the use of the true color, the normalized difference vegetation index (NDVI), Burned Area Index (BAI), and the Normalized Burn Ratio (NBR-RAW) layers as channels for data points and train them using various deep learning architectures with sigmoidal (binary) predictions for wildfire existence or lack thereof. They then discuss how such predictions can be integrated within the context of a carbon accounting system for nature-based solutions. 
\\
Several machine learning methodologies have been extensively employed for wildfire detection, as demonstrated by various studies in the literature. A significant portion of these strategies, however, have primarily relied on non-imagery data from historical wildfire incidents. For instance, statistical techniques such as Support Vector Machines (SVM) and logistic regression have found application in wildfire detection and predictive risk modeling \cite{cortez2007} \cite{nhongo2019} \cite{chang2013}.

More contemporary studies have shifted their focus to deep learning models that leverage remotely sensed satellite imagery for wildfire detection \cite{radke2019} \cite{toan2019} . A study by N. Toan et al. in \cite{toan2019} utilized the high-resolution data generated by the Geospatial Operating Environmental Satellites-16 (GEOS-16), resulting in an impressive F-1 score of 94\% on the test set. Despite the encouraging results, the generalizability of this approach is restricted by the high costs and technical complexities associated with GEOS-16 satellite data. Other researchers have attempted to tackle the wildfire detection problem using transfer learning with architectures such as Inception-V3 and ResNet-50, yielding varying degrees of success \cite{khan2019}\cite{souza2020}\cite{lin2018} .

In another work, the authors explored the application of deep learning models on Landsat-8 satellite imagery for wildfire detection \cite{pereira2021active}. The study achieved training and test accuracies of 86\% and 87\%, respectively, despite the limitations of Landsat's temporal (every 16 days) and spatial (30m - 60m) resolutions. Our study builds on this by using the Sentinel satellite's data, which is not only freely accessible like Landsat's but also offers superior temporal and spatial resolutions.

A notable concern with the methodology of \cite{lin2018} is the potential bias introduced by sourcing wildfire location data from Landsat's computational estimations. By independently obtaining wildfire location-date pairs from the US government database GEOMAC and subsequently extracting corresponding satellite images, we aim to minimize such bias in our study.

Furthermore, the study \cite{lin2018} overlooked additional potentially useful layers/bands available from satellites such as Landsat or Sentinel. In contrast, a more robust approach such as \cite{wu2020satellite} may incorporate layers such as Normalized Differential Vegetation Index (NDVI), Moisture Index (MSI), Burned Area Index (BAI), and Normalized Burn Index (NBI) from Sentinel, with the aim of developing more robust models that can be incorporated into a robust emissions liability management framework.

\subsection{From Wildfire Detection to Wildfire Emissions Quantification}
Understanding the dynamics of wildfires in the context of carbon accounting has taken on an increasingly prominent role, given the escalating frequency and intensity of such events under changing climate conditions. While the application of deep learning models for wildfire detection has advanced significantly, these models primarily serve in identifying the occurrence and location of such disturbances. Yet, to fully quantify the carbon impact of these wildfires and facilitate the assessment of emissions liabilities, we need to advance beyond detection. This entails developing computational methods capable of quantifying the amount of carbon emissions associated with detected wildfires. Here, we present an exploratory approach to extend the deep learning wildfire detection model to the quantification of associated emissions. Among others, a key benefit of these remote sensing based approaches is that they can be used even when it is not feasible to gather inventories after a large wildfire. They can also scale to several locations around the world where historical data on fire inventories before and after may not be available \cite{xu2020estimating}

\paragraph{Geo-Spatial Boundaries Estimation}
The first step in emissions estimation from wildfire incidents is accurately determining the geospatial boundaries of the fire. A detected wildfire incident can be delineated using the outputs from the wildfire detection model, where the model identifies pixels that represent wildfire. This allows us to demarcate the spatial extent of the wildfire which can then be used in subsequent steps for carbon emissions calculations.
For the geo-spatial boundaries estimation, we can use a similar CNN architecture to the base detection model such that we are also predicting for each pixel whether or not they are within a burned area. This CNN based approach to wildfire boundaries estimation has already been successfully demonstrated by Google AI researchers \cite{benhaim2023}. Instead of a separate network/model for boundaries estimation, a single model with two outputs, one for predicting whether or not there's a wildfire and another for predicting the boundaries of that wildfire could be used as is seen in the work of fellow Stanford researchers in \cite{wu2020satellite}, saving computational time and resources.

Another architecture and approach that could be used for this would be undertaking a semantic segmentation based on the U-Net architecture, which was initially designed for biomedical image segmentation but has since been used in a wide range of applications, including satellite imagery analysis \cite{ronneberger2015}. Using training data pairs of input images and corresponding target segmentation masks for wildfire boundaries, we can train a U-net model using a suitable loss function such as a pixel-wise cross-entropy loss function. During training, the model learns to classify each pixel to its corresponding class (fire or non-fire) and also captures the spatial structure and shape of the fire.

\paragraph{Vegetation Carbon Stock Assessment}
We have hitherto mentioned that the Normalized Differential Vegetation Index (NDVI) is an index that is measured and tracked by various earth observation satellites such as the Sentinel-2 among others. This index is a key parameter that can be used to measure vegetation biomass and subsequently, carbon stock. NDVI values range between -1 and +1, with higher values indicating higher vegetation density. 
Research efforts such as \cite{zhang2022estimation} and \cite{xu2020estimating} already provide a robust computational methodology for estimating aboveground biomass (ABG) from NDVI. An example of such a relationship is $AGB = a * (NDVI - b)^c$, where a, b, and c are constants derived from the data. These relationships can vary depending on the type of vegetation and the specific region. For example, in \cite{zhang2022estimation}, the used regression to estimate models for estimating aboveground biomass from NDVI for four types of vegetation, namely, arbors, shrubs, herbs and crops in a delta oasis region, with the NDVI image data coming from Landsat-8. 
Having obtained the aboveground biomass, we can estimate the carbon stock using assumptions drawn from literature on carbon fraction per kg of biomass for that location and then computing the delta pre wildfire and post wildfire. 

We can also use another approach that draws from the Burned Area Index (BAI) and Normalized Burn Ratio (NBR) channels that are available from the satellite services. In \cite{chang2016spatial}, the authors leverage historical forest inventory data to estimate pre-fire biomass density.  Building on Chang et al's  finding as reported in  \cite{chang2016spatial} that NBR is highly correlated with field-measured Composite Burn Index (R2 = 0.63, p $<$ 0.0001) a global used index for quantifying burn severity, the authors of \cite{xu2020estimating} then map various NBR values to burn severity levels and subsequently consumed foliage levels. All of this were then used to estimate the biomass consumption and carbon stock loss of a species, given a particular burn severity and for a pixel on the satellite image.

\paragraph{Quantifying the Emissions}
Carbon emissions can then be calculated as the difference between the pre- and post-wildfire carbon stocks. This measure effectively quantifies the carbon that has been released into the atmosphere as a result of the wildfire. We can multiply the estimated carbon that is emitted by the emissions factor for CO2 (using the fact that 1 kg of carbon, when fully oxidized, becomes 44/12 kg of CO2) to obtain the estimated CO2 emissions. For other gases such as methane, some studies show that CO2 equivalents of CH4 and N2O might make up around 5-10\% of the total emissions \cite{urbanski2014} \cite{sinha2003emissions}. 

\subsubsection{Sources of Uncertainty}
Understanding and mitigating uncertainty is crucial in any modeling process, and this extends to estimating wildfire emissions. 
The key sources of uncertainty in the emissions estimation approach we described are the Geo-spatial boundaries estimation, Vegetation carbon stock assessment, Emission factors, and the Burned Area Index and Normalized Burn Ratio assessments. For example, errors in boundary estimation directly influence the total area calculation for wildfires, which impacts the volume of biomass burned and therefore the emissions.
Similarly, estimating the biomass in an area involves a degree of uncertainty. This can come from the remote sensing measurements themselves, the models used to convert these measurements to biomass estimates, and the specific factors related to the vegetation type and density in the area. There is also inherent variability in emission factors due to differences in fire behavior, vegetation types, and combustion efficiency. Additionally, the proportion of different gases (CO2, CH4, N2O) can vary widely depending on the specific conditions of the fire. Finally, While the BAI and NBR indices provide useful indicators, they are proxies and may not perfectly capture the burned area or the degree of burning. They may be influenced by factors such as atmospheric conditions, sensor angle, and post-fire changes \cite{fornacca2018evaluating}.
While this work does not go into the quantification of these uncertainty, we recommend that they be quantified using uncertainty quantification techniques such as Monte-Carlo simulations, confidence intervals and sensitivity analysis as appropriate. Given the inherent uncertainties in predicting wildfires and their emissions, it could be important to perform sensitivity analyses and incorporate margins of safety. For instance, one could calculate the adjusted carbon sequestration under different scenarios of wildfire probability and emissions, to understand the potential range of outcomes.  Then such uncertainties may potentially be mitigated by exploring multiple data sources and models, improving measurements, and exploring techniques such as the Value of Information (VoI) to determine whether additional data would reduce uncertainty \cite{kochenderfer2022algorithms}.

\subsection{Incorporation of Wildfire Risk in Emissions Liability Management}
We have discussed how aforementioned wildfire impact and associated emissions estimation framework can significantly aid large-scale leakage-aware carbon accounting for vegetation based projects. Such a framework can also be extended into wildfire risk modeling, leading to more risk-aware decision-making for emissions liability management practices. Discussed as follows, are two such ways in which this extension can be made:
\begin{itemize}
    \item \textbf{Crediting for Reforestation Carbon Sequestration:}
In addition to identifying sources of uncertainty in our wildfire emissions estimation, it is also important that wildfire risks are incorporated when such estimates are used for important decisions such as carbon crediting.  Incorporating wildfire risk into afforestation carbon accounting involves understanding both the probability of wildfires occurring in a given region and the potential magnitude of carbon emissions if such wildfires occur. This is an essential consideration because forests that have significant carbon sequestration potential may also have high wildfire risks, which can lead to large carbon emissions.

To incorporate wildfire risk, we can modify the carbon accounting methodology for afforestation projects by introducing a risk factor derived from the probability of wildfires and their potential emissions. In particular, we can achieve this by first estimating the probability of wildfires occurring in the afforestation area over a specific time horizon. This could be based on historical wildfire data and models that consider factors like climate, vegetation, and human activities. This kind of approach has already been undertaken in prior research as we see in \cite{preisler2004probability} and \cite{finney2011simulation} where the authors generated wildfire risk using historical data. 

For our own analysis, we can specify the probability of wildfire occurring over a given time horizon in as P(W), where W stands for wildfire.
Subsequently, we estimate the potential carbon emissions if a wildfire were to occur. This estimation could leverage the method we have outlined earlier, based on estimates of vegetation carbon stocks and burn severity.
We can similarly denote this as $E(W)$, where E stands for emissions.

We can then compute the wildfire risk factor as the expected carbon emissions from wildfires as the product of the wildfire probability and the estimated potential emissions. This can be represented as follows: \\

\textbf{Expected emissions due to wildfire: } \\
\begin{align*}
 E_{\text{expected}} = P(W) \times E(W)\\
\end{align*}

Finally, we can adjust the carbon accounting for the afforestation project by subtracting the expected emissions due to wildfire risk from the estimated carbon sequestration. \\

\textbf{Adjusted Carbon Sequestration: } \\
\begin{align*}
 S_{\text{adjusted}} = S_{\text{estimated}} - E_{\text{expected}}
\end{align*}
\\

This adjusted measure of carbon sequestration would provide a more realistic estimate of the net carbon benefits of the afforestation project, taking into account the wildfire risk. In particular, we are better able to account for the potential loss of carbon stocks due to wildfires, and now have a more accurate and risk-aware measure of the carbon sequestration potential of afforestation projects. The approach would also help to guide decisions about where to locate these projects to maximize their net carbon benefits, taking into account both the potential for carbon sequestration and the risk of wildfires.

\item \textbf{Insurance of Reforestation Projects}  
Traditionally, buffer pools have been used as a way to manage leakage risk in reforestation projects. The existence of such pools guarantee that carbon offset credits are still provided even when trees are destroyed via wildfires. However, in locations such as California where this concept has been championed, there has been an increasing depletion of the buffer pools as the frequency of wildfires increase. Alternative risk-aware insurance mechanisms are needed by insurance providers not only for reforestation projects but also for property located in these increasingly wildfire prone areas. In the absence of a sufficient way to model wildfire risk in their underwriting processes, insurance companies are increasingly abstaining from insuring reforestation projects \cite{greenly2023}. The approach we have proposed can allow insurance to model these risks to inform their underwriting decisions, allowing them to strike a balance between risk and returns. These insurance companies can also then limit their exposure in high-risk areas while insuring those in low-risk regions and perhaps prompt policyholders to implement specific risk mitigation practices. 
\\

Another application of AI-driven wildfire risk modeling to the insurance of reforestation projects is the estimation of Incurred But Not Reported (IBNR) Reserves. IBNR liabilities refer to claims or liabilities faced by an insurance company that are expected to have occurred but have not yet been reported to the insurance company \cite{geiger2022analysis}. The notion of IBNR arises from the fact that in reality, there is often a lag time between when a claim-generating event occurs and when it is actually occurs. Thus, as insurance companies estimate their future liabilities and plan financial reserves to cover them, they need to also take into account these types of liabilities. Traditionally, the estimation of IBNR reserves has been a challenging task that often entails the combination of actuarial judgment and statistical analysis and has been subject to a lot of criticisms in its limited accuracy \cite{norberg1986contribution}. The wildfire risk modeling framework discussed previously can support insurers in more robustly and accurately. 

\end{itemize}

\begin{figure*}[t!]
\centering
\includegraphics[width=\textwidth]{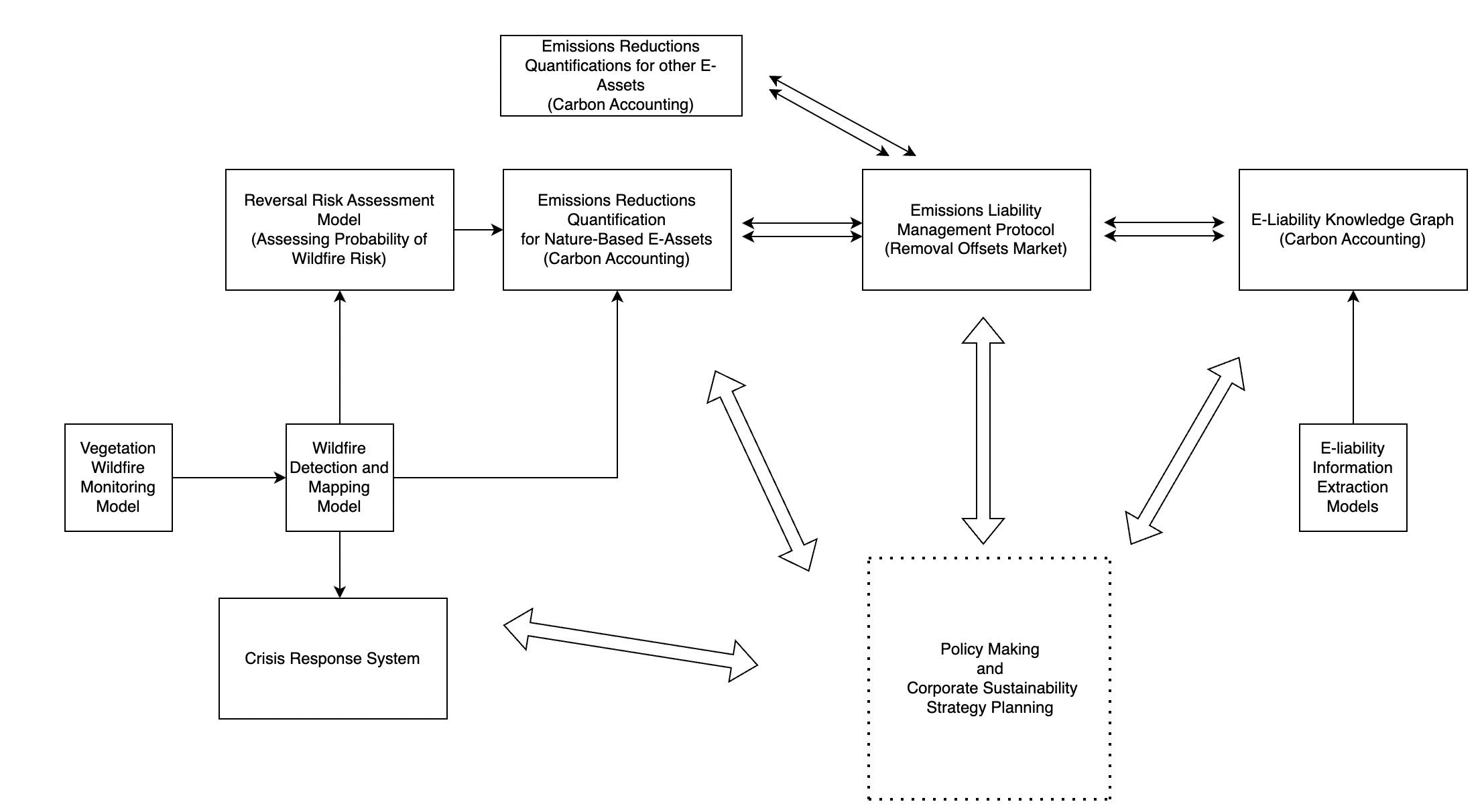}
\caption{Integrated System for Emissions Liability Management Decision Support}
\label{dynamicfr}
\end{figure*} 

\subsection{Overall AI-aided Framework for Nature-Based Solutions}
The approach to wildfire emissions estimation that we have delineated is part of a broader, data-driven strategy towards carbon accounting in the context of nature-based climate solutions, including afforestation. Given the significant role that forests play in carbon sequestration, accurately accounting for carbon stocks and fluxes in these ecosystems is crucial for assessing the effectiveness of these climate solutions and informing policy and management decisions.
We have demonstrated how such a machine learning driven approach could aid real-time carbon stock and carbon flux accounting, enhancing broader efforts towards the mitigation of climate change. They can be integrated to be part of an overarching computational decision support system for nature based climate solutions which performs the following functions:
\begin{itemize}
\item \textbf{Real-time Monitoring of Forest and Agricultural Conditions:} With satellite data being continuously updated, deep learning models can analyze the most recent data to track changes in forest conditions in near-real-time. This provides valuable data on changes in forest density, vegetation health, and changes in land use, which can inform the carbon accounting process and guide management decisions as we have previously illustrated.
\item \textbf{Dynamic Carbon Accounting:} Traditional carbon accounting methods often rely on periodic inventory data, which can be out of date and fail to capture recent changes in forest conditions. An AI-aided system can provide dynamic carbon accounting, updating carbon stock and flux estimates as new satellite data becomes available. This approach allows for a more accurate and up-to-date picture of the carbon balance of a forest. 

\item \textbf{Improved Emissions Liability Management:} If a nature-based solution is being used as part of an offset program to manage emissions liability, then adequate monitoring, verification and incorporation of reversal risk is essential. The monitoring and emissions estimation approaches we have presented can be combined with a wildfire risk engine to robustly capture the expected carbon sequestered for nature based solutions. This can then be used to appropriately provide carbon credits for a vegetation based offset project. 

\item \textbf{Early Wildfire Detection and Response:} As we have explored in detail, machine learning models can be used to identify wildfires at their early stages, even before they are detected by traditional means. This allows for more rapid response to these events, potentially reducing their severity and the associated carbon emissions.
\item \textbf{Risk Assessment and Mitigation Planning:} Machine learning models can also help to identify areas at high risk of wildfires or other disturbances, by analyzing patterns in the environmental variables associated with these events. This information can guide mitigation efforts, such as fuel reduction or controlled burns, to reduce the risk of severe wildfires and the associated carbon emissions.
\item \textbf{Policy and Management Decision Support:} By providing a more accurate and timely picture of forest conditions and carbon balance, and by identifying areas of risk, an AI-aided system can support policy decisions and forest management practices. For example, it could inform decisions about where to focus afforestation efforts, or how to allocate resources for fire prevention and response.
\end{itemize}

Figure \ref{dynamicfr} illustrates how such a comprehensive decision support system could function. 


\section{Conclusion and Future Directions}

This paper began with a discussion, of the disparate nature AI-based emissions monitoring viz. a viz carbon accounting practices, the management of emissions and associated reversal risk, and overall climate mitigation policies. After discussing some of these elements, we then proposed an integrative computational framework that leverages AI to provide comprehensive climate decision support for nature-based solutions. 
Decision-makers can use the framework developed to provide comprehensive computational decision support for nature-based climate solutions. We have discussed how an element of this framework can address real-time monitoring of forest and agricultural conditions especially when there are wildfires using satellite imagery and computer vision. We can also integrate this with a dynamic vegetation carbon stock assessment to automatically estimate emissions and reversals due to these wildfires. These can then ultimately inform adjustments to carbon sequestration planning, including the insurance of newer offsets. 
Ultimately, by integrating these functionalities, an AI-aided decision support system could greatly enhance our ability to manage and protect our forests, maximize their carbon sequestration potential, and mitigate the impacts of wildfires and other disturbances. This represents an exciting frontier for the application of AI in support of nature-based climate solutions.

\bibliographystyle{aaai} \bibliography{references.bib}

\end{document}